\documentclass{article}

\PassOptionsToPackage{numbers,sort&compress}{natbib}
\usepackage[preprint]{neurips_2024}
\usepackage[colorlinks=true,linkcolor=blue,citecolor=blue]{hyperref}

\makeatletter
\renewcommand{\@noticestring}{Preprint Version. Under review.} % 只保留 Preprint
\makeatother   % ← 加上这行

\usepackage[utf8]{inputenc} % allow utf-8 input
\usepackage[T1]{fontenc}    % use 8-bit T1 fonts
\usepackage{hyperref}       % hyperlinks
\usepackage{url}            % simple URL typesetting
\usepackage{booktabs}       % professional-quality tables
\usepackage{amsfonts}       % blackboard math symbols
\usepackage{graphicx}       % for including graphics
\usepackage{nicefrac}       % compact symbols for 1/2, etc.
\usepackage{microtype}      % microtypography
\usepackage{xcolor}         % colors
\usepackage{subcaption}
\usepackage{graphicx}
\usepackage{float}
\usepackage{algorithm}
\usepackage{algpseudocode}
\usepackage{amsmath}
\usepackage[numbers]{natbib}

\title{TaoCache: Structure-Maintained Video Generation Acceleration}

\author{%
  Zhentao Fan\thanks{Corresponding author. } \\
  Huawei Inc.\\
  \texttt{zhentao.fan@mail.utoronto.ca} \\
  \And
  Zongzuo Wang \\
  Huawei Inc.\\
  \And
  Weiwei Zhang \\
  Huawei Inc.\\
}

\begin{document}

\maketitle

\begin{abstract}
Existing cache-based acceleration methods for video diffusion models primarily skip early or mid denoising steps, which often leads to structural discrepancies relative to full-timestep generation and can hinder instruction following and character consistency. We present TaoCache, a training-free, plug-and-play caching strategy that, instead of residual-based caching, adopts a fixed-point perspective to predict the model’s noise output and is specifically effective in late denoising stages. By calibrating cosine similarities and norm ratios of consecutive noise deltas, TaoCache preserves high-resolution structure while enabling aggressive skipping. The approach is orthogonal to complementary accelerations such as Pyramid Attention Broadcast (PAB) and TeaCache, and it integrates seamlessly into DiT-based frameworks. Across Latte-1, OpenSora-Plan v110, and Wan2.1, TaoCache attains substantially higher visual quality (LPIPS, SSIM, PSNR) than prior caching methods under the same speedups. 
\end{abstract}

\section{Introduction}

Diffusion models have recently shown remarkable capability in high-quality video generation, particularly with Diffusion Transformers (DiTs)~\cite{peebles2023dit, latte}. Despite state-of-the-art fidelity, their iterative denoising inherently incurs heavy computation: producing high-resolution or long-duration videos typically requires hundreds of sequential inference steps, which limits real-time and interactive use.

To reduce this cost without retraining, recent work explores \emph{caching} strategies that reuse intermediate outputs across timesteps. \textsc{AdaCache}~\cite{adacache} dynamically selects timesteps for recomputation via a feature-distance metric and motion regularization, allocating inference adaptively to video content. \textsc{TeaCache}~\cite{teacache} leverages timestep embeddings to predict output variation and skips steps whose predicted residuals fall below calibrated thresholds. \textsc{MagCache}~\cite{magcache} further simplifies the criterion using residual-norm magnitude, based on the observation that residual-norm ratios decrease through denoising. However, these approaches primarily skip early or mid stages; the resulting small discrepancies can compound and manifest later as degraded spatial structure and weakened high-frequency details—precisely where late-stage denoising is visually critical.

We address these limitations with \textbf{TaoCache}, a training-free, plug-and-play mechanism tailored to effective caching in the late denoising stages. Rather than relying on first-order residual approximations, TaoCache adopts a fixed-point view of the model’s noise prediction and explicitly models \emph{second-order} noise deltas. By calibrating norm ratios and cosine similarities from consecutive late-stage steps, TaoCache predicts the model outputs for skipped timesteps while preserving global geometric consistency—even under aggressive skipping at high resolutions. The method introduces only a single lightweight calibration step and integrates seamlessly with DiT-based frameworks.

We evaluate TaoCache across diverse video generation stacks, including Latte-1 2B, OpenSora-Plan v110, and Wan2.1-1.3B. Under matched speedups, TaoCache delivers higher visual quality—measured by LPIPS, SSIM, and PSNR—than prior caching methods. Moreover, it complements orthogonal accelerations such as TeaCache and Pyramid Attention Broadcast (PAB), further improving end-to-end efficiency.

\section{Related Work}

\subsection{Diffusion Models for Video Generation}
Diffusion probabilistic models~\cite{ho2020ddpm} have become the leading paradigm for high-quality video generation, surpassing GAN-based and autoregressive approaches in visual fidelity and training stability. Architectures have progressed from UNet-style designs (e.g., Make-A-Video~\cite{singer2022makeascope}) to Diffusion Transformers (DiTs)~\cite{peebles2023dit, latte}. State-of-the-art systems such as Open-Sora-Plan and Wan2.1 produce compelling results at resolutions from various scales. However, the large number of inference steps required still hinders real-time and interactive applications, motivating substantial work on inference acceleration.

\subsection{Acceleration Methods for Diffusion Inference}
\paragraph{Step reduction and advanced samplers.}
Step-reduction methods include training-based approaches such as progressive distillation~\cite{sauer2023kdpm}, sampling pseudo-knowledge distillation~\cite{bao2023spkd}, and distribution matching~\cite{yin2024efficient}. Training-free ODE/ SDE solvers (e.g., DPM-Solver~\cite{dpmsolver2022} and UniPC~\cite{zhao2023unipc}) also accelerate sampling substantially. In practice, training-based methods require model retraining, while training-free solvers can face stability or quality trade-offs at very low step counts, especially under guidance, limiting direct applicability to video DiTs.

\paragraph{Spatial and temporal sparsity.}
Computational sparsity reduces token counts or focuses compute on salient regions. Token merging~\cite{bolya2023tokenprune} merges redundant spatial tokens; region-adaptive sampling~\cite{nitzan2024region} concentrates resources where motion is present. Pyramid Attention Broadcast (PAB)~\cite{pab2024} hierarchically reuses multi-scale context, and Sparse VideoGen~\cite{videogen2025} leverages temporal attention sparsity. These techniques are orthogonal to timestep skipping and can complement caching.

\subsection{Feature Caching Strategies}
Feature-level caching reuses intermediate activations to avoid redundant computation without retraining. \textsc{AdaCache}~\cite{adacache} selects recomputation steps using a feature-space distance with motion regularization, adapting compute to content. \textsc{TeaCache}~\cite{teacache} predicts output variation from timestep embeddings and skips steps whose calibrated residual estimates fall below a threshold. \textsc{MagCache}~\cite{magcache} uses a magnitude-based rule driven by the (near-)decay of residual-norm ratios to enable global skipping after minimal calibration. \textsc{Skip-DiT}~\cite{skipdit} inserts long-skip connections so deep features evolve slowly, enabling extensive reuse at the cost of fine-tuning. \textsc{DuCa}~\cite{duca} interleaves aggressive layer caching with periodic token-wise recomputation to heal accumulated error, though additional forward passes can dilute net speedup.

\subsection{Content-Adaptive Computation}
Several methods tailor computation to content characteristics, e.g., spatial activity or motion. Region-based strategies refine active regions~\cite{nitzan2024region}, and \textsc{AdaCache} adjusts allocation using optical flow~\cite{adacache}. TaoCache is complementary: it operates primarily along the timestep dimension and integrates with existing spatially adaptive techniques.

\subsection{Positioning TaoCache}
TaoCache frames feature caching as a fixed-point estimation problem over \emph{second-order} noise deltas, targeting the visually critical late denoising stages. Unlike first-order approximations, it preserves geometric and structural fidelity at high resolutions, requires no model modification, and needs only lightweight calibration. Experiments show that TaoCache outperforms prior caching baselines under matched speedups and pairs well with orthogonal accelerations in later denoising.

\section{Methodology}

\subsection{Observation}

\begin{figure}[htbp]
  \captionsetup[subfigure]{margin=3pt,justification=centering}
  \centering
  \begin{subfigure}[b]{0.36\textwidth}
    \centering
    \includegraphics[width=\linewidth]{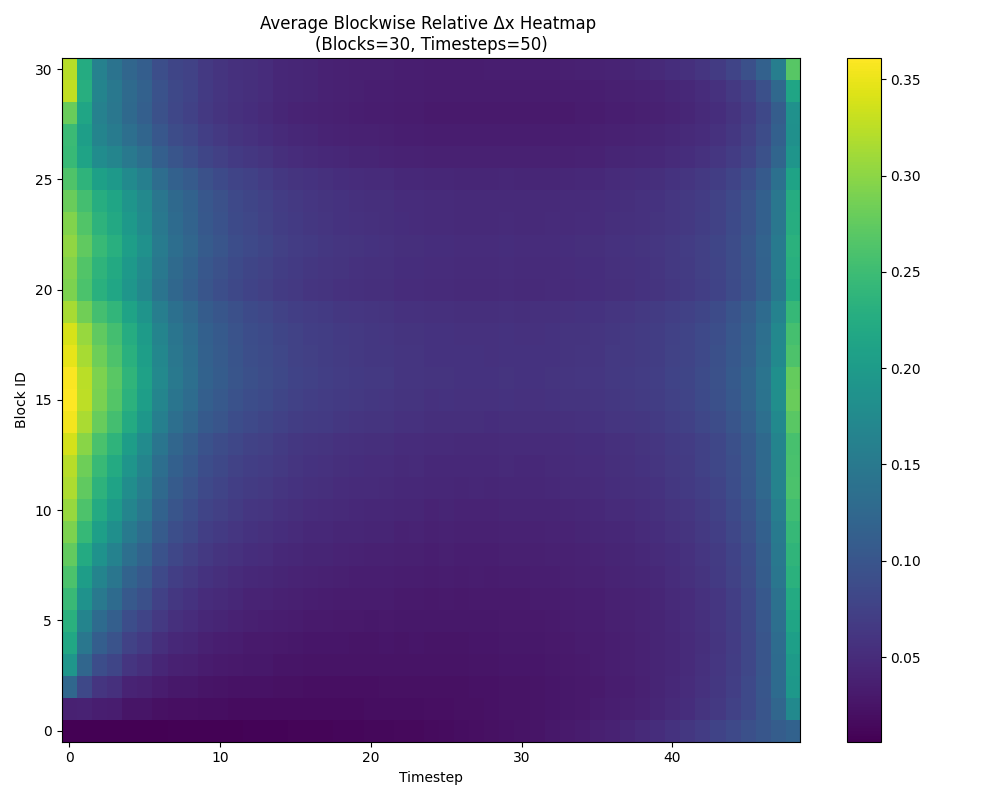}
    \caption{Blockwise relative \emph{input} delta (first-order) heatmap in Wan2.1-1.3B. The $31$-st row corresponds to the last layer’s output.}
    \label{fig:obs-heatmap}
  \end{subfigure}\hfill
  \begin{subfigure}[b]{0.315\textwidth}
    \centering
    \includegraphics[width=\linewidth]{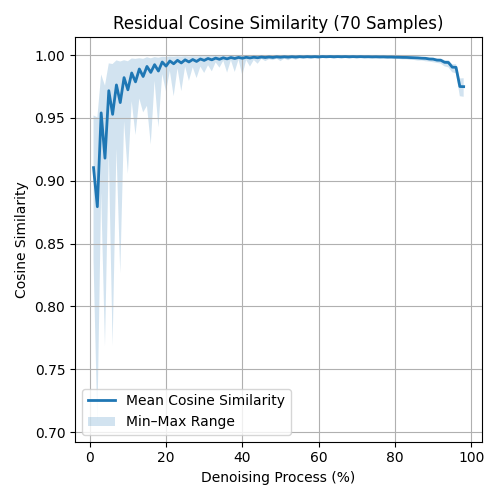}
    \caption{Timestep-wise cosine similarity of single-step residuals in Wan2.1-1.3B.}
    \label{fig:obs-res-cos}
  \end{subfigure}\hfill
  \begin{subfigure}[b]{0.315\textwidth}
    \centering
    \includegraphics[width=\linewidth]{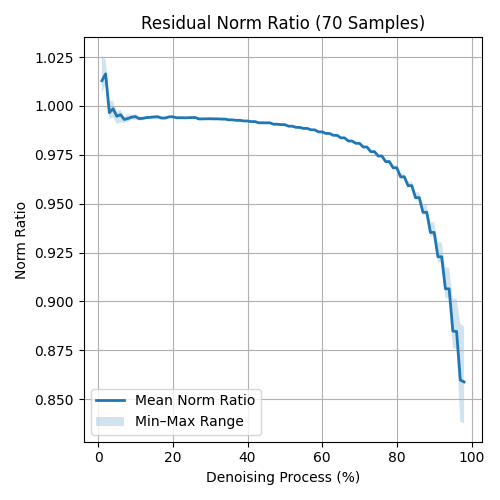}
    \caption{Timestep-wise norm ratio of single-step residuals in Wan2.1-1.3B.}
    \label{fig:obs-res-norm}
  \end{subfigure}
  \caption{Layerwise/stepwise statistics that inform caching policy. Mid-range steps show smaller first-order changes (a), while early/late steps exhibit higher variance in residual cosine similarity (b) and norm ratios (c), making fixed thresholds less reliable across prompts.}
  \label{fig:obs}
\end{figure}

Different choices of \emph{what} to cache naturally induce different caching policies. 
As shown in Fig.~\ref{fig:obs}\,(a), the \emph{blockwise relative input/output delta} is smaller in the mid-range denoising steps, which explains why many feature-caching methods preferentially skip there. For approaches such as \textsc{TeaCache} and \textsc{MagCache} that rely on single-step residual signals—distance between the input and output of a one-step forward DiT—their cosine similarity and magnitude (norm) ratios display larger variance in the early and late stages (Fig.~\ref{fig:obs}\,(b,c)). Since these metrics are \emph{a priori} unknown for a specific prompt, static or globally calibrated thresholds can lead to unstable skip allocations across timesteps.

\begin{figure}[H]
  \centering
  \includegraphics[width=1\linewidth]{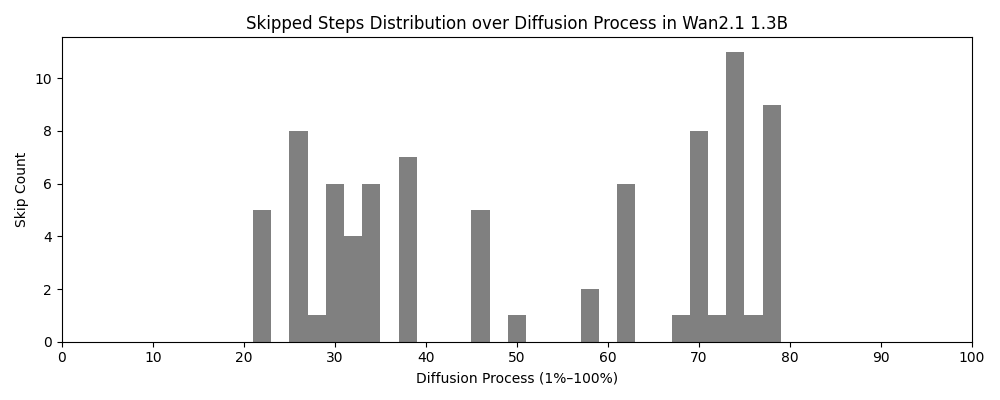}
  \caption{\textsc{TeaCache} skip counts across denoising steps under total skip budgets of $2\%,4\%,6\%,\ldots,24\%$. Early/late stages are harder to skip consistently due to the higher variance of first-order residual signals.}
  \label{fig:tea-skip}
\end{figure}

Moreover, early steps are crucial for latent denoising~\cite{magcache}; errors from skipping there can \emph{propagate} and manifest as structural drift. While later steps may recover low-frequency fidelity, the damage to \textbf{instruction following} and \textbf{character consistency} is harder to repair.

To seek a signal that is stable \emph{at late timesteps} yet light-weight to compute and store, we examine \emph{output-noise deltas}:
\[
\Delta \boldsymbol{\epsilon}_t \;:=\; \boldsymbol{\epsilon}_\theta(\mathbf{x}_t, t)\;-\;\boldsymbol{\epsilon}_\theta(\mathbf{x}_{t+1}, t{+}1),
\]
and measure their cosine similarity and norm ratio across consecutive late-stage steps:
\[
\mathrm{cos\_sim}(t)=\frac{\langle \Delta \boldsymbol{\epsilon}_{t}, \Delta \boldsymbol{\epsilon}_{t+1}\rangle}{\|\Delta \boldsymbol{\epsilon}_{t}\|\,\|\Delta \boldsymbol{\epsilon}_{t+1}\|}, 
\qquad
\mathrm{norm\_ratio}(t)=\frac{\|\Delta \boldsymbol{\epsilon}_{t+1}\|}{\|\Delta \boldsymbol{\epsilon}_{t}\|}.
\]

\begin{figure}[H]
  \centering
  \includegraphics[width=1\linewidth]{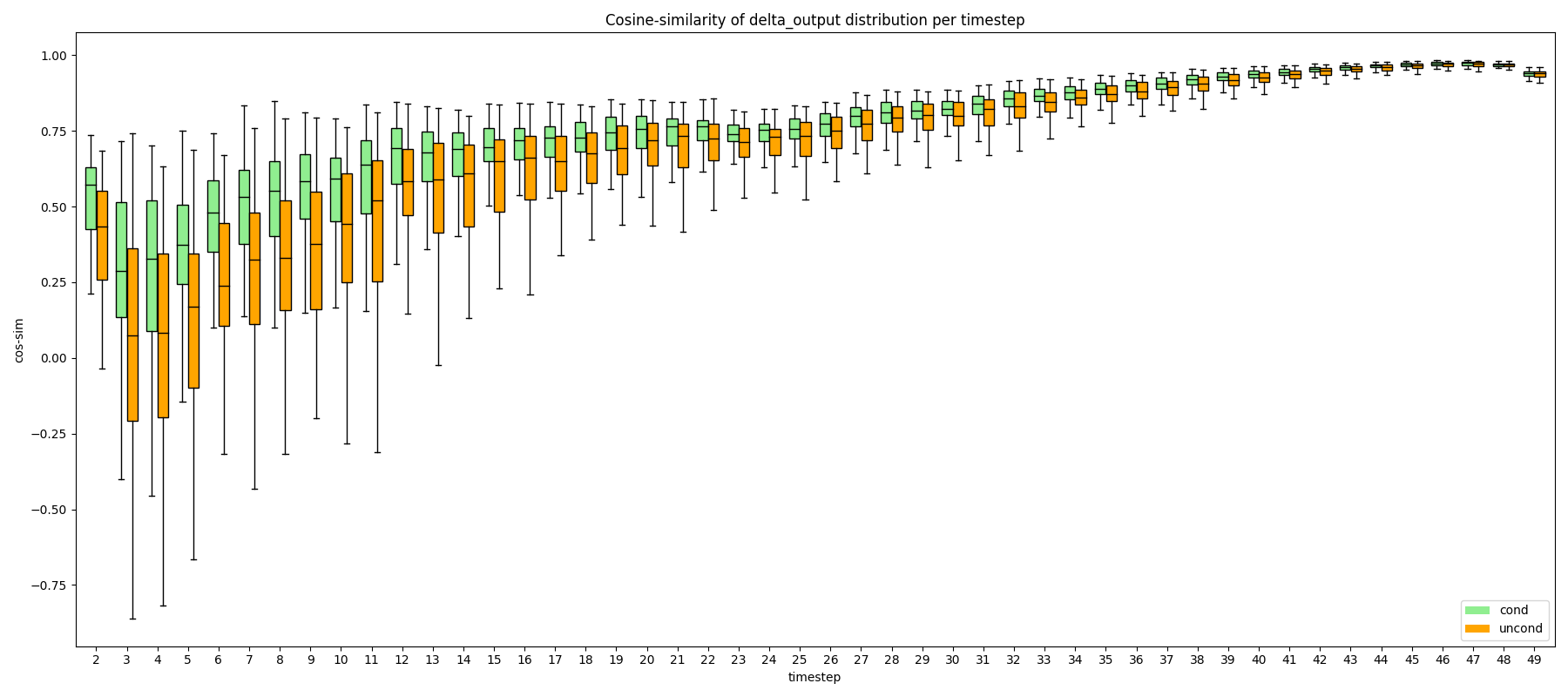}
  \caption{Cosine similarity of \emph{output-noise deltas} in Wan2.1-1.3B (``\textit{cond}'' denotes the positive-conditioning stream; ``\textit{uncond}'' the negative). Late-stage values converge and remain highly correlated.}
  \label{fig:delta-cos}
\end{figure}

\begin{figure}[H]
  \centering
  \includegraphics[width=1\linewidth]{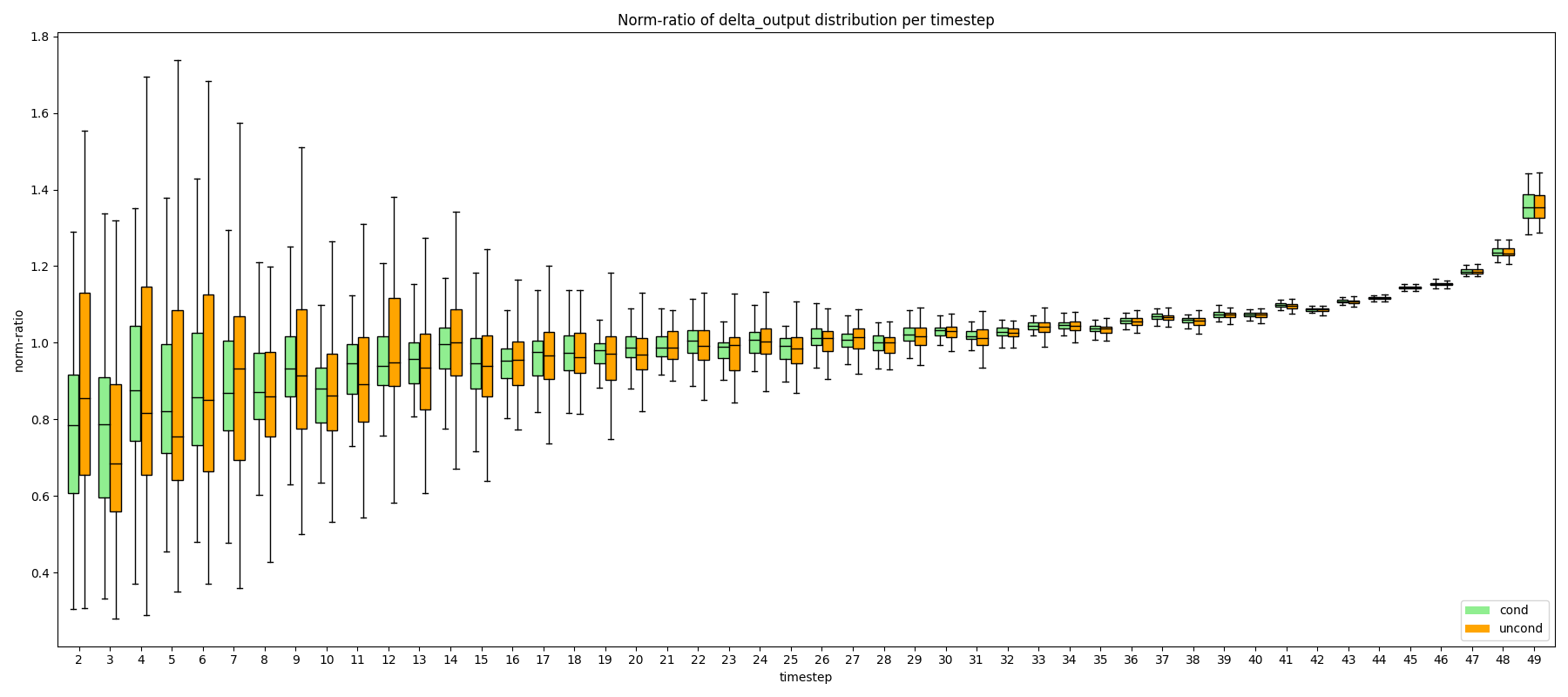}
  \caption{Norm ratio of \emph{output-noise deltas} in Wan2.1-1.3B (cond/uncond as above). Ratios stabilize in late denoising, yielding a predictable scale relationship.}
  \label{fig:delta-norm}
\end{figure}

Empirically (Fig.~\ref{fig:delta-cos}--\ref{fig:delta-norm}), both $\mathrm{cos\_sim}(t)$ and $\mathrm{norm\_ratio}(t)$ \emph{converge} during late denoising for both conditional and unconditional streams. This property provides a simple, robust late-stage cache signal that does not require heavy, layer-wise state. Leveraging it enables a caching policy that maintains global structure while remaining lightweight and training-free—motivating the fixed-point, second-order design of \textbf{TaoCache}.

\subsection{Design}

\paragraph{Fixed-point view of output-related caching.}
At late denoising steps, the DiT forward map is close to an identity transform over small neighborhoods of the latent trajectory. 
Let $f$ denote a generic one-step map; a classical fixed-point intuition is
\[
  y + \delta y = f(y), 
  \qquad 
  f(y + \delta y) \;\approx\; (y + \delta y) + \delta y,
\]
i.e., consecutive increments are small and strongly correlated. 
For video DiTs, we apply this view to the \emph{predicted noise}.

\paragraph{Notation.}
Let $\hat{\epsilon}_t := \epsilon_\theta(\mathbf{x}_t, t)$ be the model’s noise prediction at timestep $t$ for latent $\mathbf{x}_t$, and $\mathrm{SchedulerStep}$ the sampling update (Euler, DPM-Solver, UniPC, etc.). 
We define the \emph{output-noise delta}
\[
  \Delta_t \;:=\; \hat{\epsilon}_t - \hat{\epsilon}_{t+1}.
\]
Empirically (Fig.~\ref{fig:delta-cos}--\ref{fig:delta-norm}), in late stages the \emph{direction} of $\Delta_t$ changes slowly and the \emph{scale} evolves smoothly. 
Hence we model a second-order relation
\[
  \Delta_t \;\approx\; r_t\, \Delta_{t+1}, 
  \qquad r_t>0,
\]
where $r_t$ is a scalar \emph{norm ratio} and the direction agreement is quantified by the cosine similarity $c_t \approx 1$ between $\Delta_t$ and $\Delta_{t+1}$. 
This “scalar–times–previous-delta’’ approximation is the core of \textsc{TaoCache}.

\paragraph{Layer/solver abstraction.}
Writing the DiT as a composition $F = F_L \circ \cdots \circ F_1$ (self-/cross-attn, MLP, norms) and the sampler as $G_t(\mathbf{x}_t, \hat{\epsilon}_t)$, a single step is
\[
  \hat{\epsilon}_{t} \;=\; \epsilon_\theta(\mathbf{x}_t, t), 
  \qquad 
  \mathbf{x}_{t-1} \;=\; G_t(\mathbf{x}_t, \hat{\epsilon}_t).
\]
Because $\mathbf{x}_{t}$ and $\mathbf{x}_{t+1}$ are close in late steps, $\hat{\epsilon}_{t}$ and $\hat{\epsilon}_{t+1}$—and hence $\Delta_t$ and $\Delta_{t+1}$—exhibit high correlation, enabling the above second-order approximation without touching internal layer states.

\begin{figure}[H]
    \centering
    \includegraphics[width=0.9\linewidth]{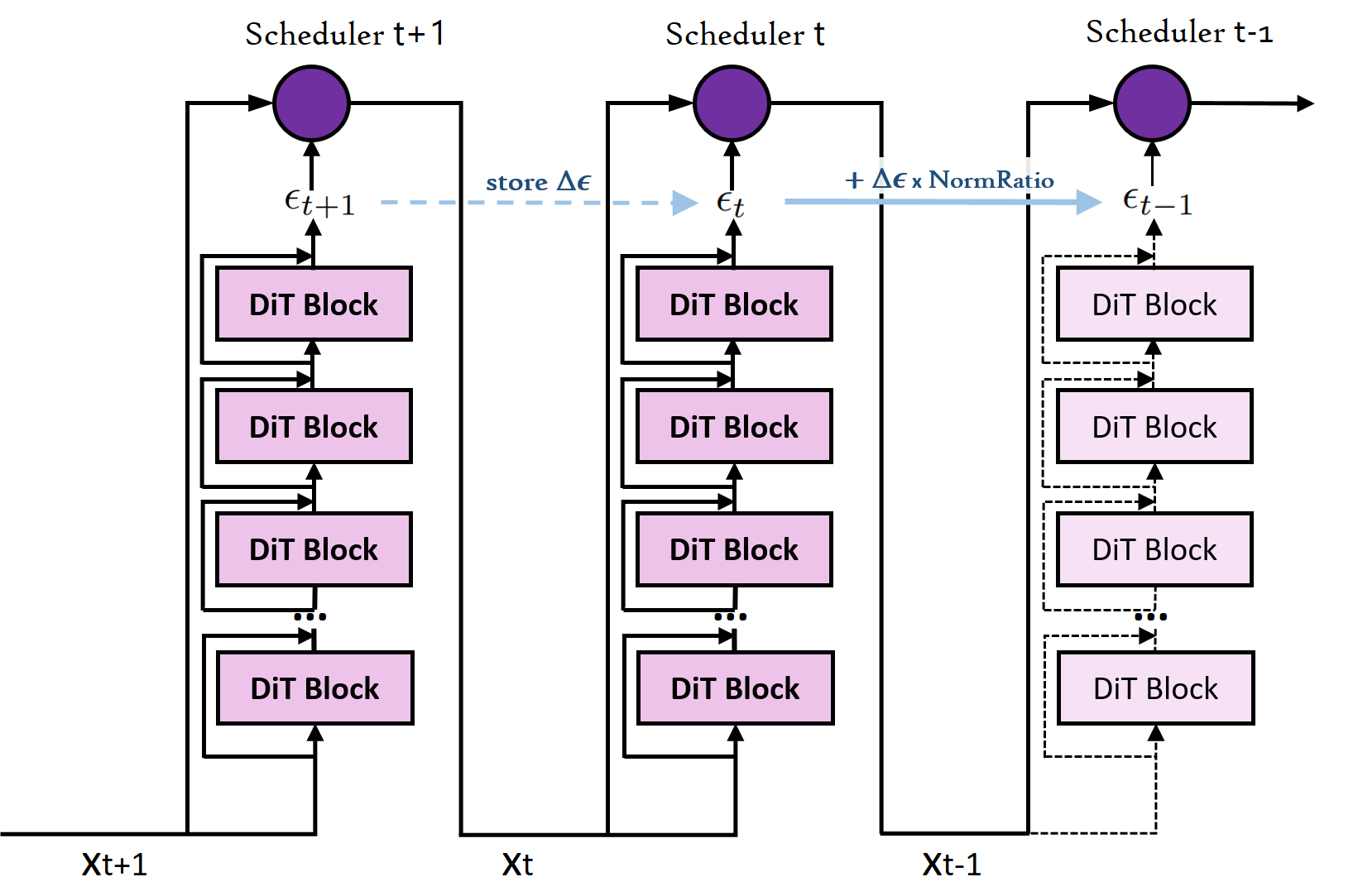}
    \caption{\textbf{TaoCache overview.} Instead of predicting $\hat{\epsilon}_t$ directly, we predict the change $\Delta_t$ from $\Delta_{t+1}$ using calibrated late-stage statistics (norm ratio $r_t$ and cosine $c_t$), then recover $\hat{\epsilon}_t = \hat{\epsilon}_{t+1} + \Delta_t$ and call the scheduler.}
    \label{fig:taocache-overview}
\end{figure}

\paragraph{One-time warmup calibration.}
For a new checkpoint, we run a small set of prompts $\mathcal{P}$ once without acceleration. 
From each trajectory we record, for every valid $t$, 
\[
  c_t^{(p)} \;=\; \frac{\langle \Delta_{t}^{(p)}, \Delta_{t+1}^{(p)}\rangle}{\|\Delta_{t}^{(p)}\|\,\|\Delta_{t+1}^{(p)}\|} 
  %\cos\!\angle\!\bigl(\Delta_t^{(p)}, \Delta_{t+1}^{(p)}\bigr),
  \qquad
  r_t^{(p)} \;=\; \frac{\|\Delta_t^{(p)}\|}{\|\Delta_{t+1}^{(p)}\|}.
\]
We then build lookup tables with both \emph{mean} and \emph{dispersion}:
\[
  C_{\text{cos}}[t] = \mathrm{mean}_p\, c_t^{(p)}, \quad 
  S_{\text{cos}}[t] = \mathrm{std}_p\, c_t^{(p)};
  \qquad
  C_{\text{ratio}}[t] = \mathrm{mean}_p\, r_t^{(p)}, \quad 
  S_{\text{ratio}}[t] = \mathrm{std}_p\, r_t^{(p)}.
\]
% (We also find the coefficient of variation $\mathrm{CV}_{\text{ratio}}[t]=S_{\text{ratio}}[t]/C_{\text{ratio}}[t]$ for stability.)

\paragraph{Deviation-aware window selection.}
Given a target skip budget $N_{\text{skip}}$, we choose a \emph{contiguous late-stage window} $W$ of length $N_{\text{skip}}$ either \textbf{manually} or by maximizing a variance-penalized score for automated skipping:
\[
  W^\star \;=\; \arg\max_W \Bigl(
      \underbrace{\mathrm{mean}_{t\in W} C_{\text{cos}}[t]}_{\text{directional agreement}}
      \;-\; \lambda \underbrace{\mathrm{mean}_{t\in W} S_{\text{cos}}[t]}_{\text{directional variability}}
      \;-\; \gamma \underbrace{\mathrm{mean}_{t\in W} S_{\text{ratio}}[t]}_{\text{scale variability}}
  \Bigr),
\]
subject to $W$ lying in the late denoising regime (enforced by an upper-bound $t$ or by requiring $C_{\text{cos}}[t]\ge\tau_{\cos}$). 
This “max-cos with deviation penalty’’ makes the skip region both \emph{predictable} and \emph{stable} across prompts.

\paragraph{Delta prediction during skipping.}
When $t\in W^\star$ (skip), we set
\[
  \widetilde{\Delta}_t \;=\; C_{\text{ratio}}[t] \cdot \widetilde{\Delta}_{t+1},
  \qquad
  \widetilde{\hat{\epsilon}}_t \;=\; \widetilde{\hat{\epsilon}}_{t+1} + \widetilde{\Delta}_t,
\]
and fall back to a \emph{refresh} (full model call) every $K$ skips (to bound drift, $K$ can be large). 
Finally, we advance the sampler via $\mathbf{x}_{t-1} = \mathrm{SchedulerStep}(\mathbf{x}_t, \widetilde{\hat{\epsilon}}_t, t)$.

\begin{algorithm}[H]
  \caption{TaoCache: Delta–Noise Calibration and Deviation-Aware Inference for DiT}
  \label{alg:taocache_delta_calib}
  \begin{algorithmic}[1]
    \Function{Calibrate}{$\mathcal{P}, \epsilon_\theta$}
      \State \textbf{Initialize:} $c[t, p]\!\gets\!0,\; r[t,p]\!\gets\!0$ for $t=T\!-\!2,\dots,1$; \Comment{means over prompts}
      \ForAll{$p \in \mathcal{P}$}
        \State $\mathbf{x}_T \gets \Call{InitNoise}{p}$
        \For{$t = T,\,T\!-\!1,\,\dots,\,1$}
          \State $\hat{\epsilon}_t \gets \epsilon_\theta(\mathbf{x}_t, t)$
          \State $\mathbf{x}_{t-1} \gets \Call{SchedulerStep}{\mathbf{x}_t,\hat{\epsilon}_t,t}$
          \If{$t \le T-1$}
            \State $\Delta_t \gets \hat{\epsilon}_t - \hat{\epsilon}_{t+1}$ \Comment{output-noise delta}
          \EndIf
          \If{$t \le T-2$}
            \State $c[t,p] = \Call{CosineSimilarity}{\Delta_t,\Delta_{t+1}}$
            \State $r[t,p] = \|\Delta_t\| / \|\Delta_{t+1}\|$
          \EndIf
        \EndFor
      \EndFor
      \State \textbf{Normalization:} $C_{\text{cos}}[t] \gets \Call{mean}{c[t]} $,\quad $C_{\text{ratio}}[t] \gets \Call{mean}{r[t]}$ \textbf{for each} $t$
      \State \textbf{Deviation:} $S_{\text{cos}}[t] \gets \Call{std}{c[t]} $,\quad $S_{\text{ratio}}[t] \gets \Call{std}{r[t]}$ \textbf{for each} $t$
      \State \Return $C_{\text{cos}}, C_{\text{ratio}}, S_{\text{cos}}, S_{\text{ratio}}$
    \EndFunction
    \\
    \Function{TaoCacheForward}{$\mathbf{x}_T, \epsilon_\theta, C_{\text{cos}}, C_{\text{ratio}}, S_{\text{cos}}, \mathrm{S}_{\text{ratio}}, N_{\text{skip}}, K$}
      \State $\mathcal{T}_{\text{skip}} \gets \Call{MaxCosSlidingWindow}{C_{\text{cos}}, N_{\text{skip}}, S_{\text{cos}}, S_{\text{ratio}}, K}$ or Manually
      % \State $\text{last\_refresh} \gets T$ \Comment{to enforce periodic refresh every $K$ steps}
      \For{$t = T,\,T\!-\!1,\,\dots,\,1$}
        \If{$(t \notin \mathcal{T}_{\text{skip}})$}
          \State $\hat{\epsilon}_t \gets \epsilon_\theta(\mathbf{x}_t, t)$
          \If{$t \le T-1$} \State $\Delta_t \gets \hat{\epsilon}_t - \hat{\epsilon}_{t+1}$ \EndIf
          % \State $\text{last\_refresh} \gets t$
        \Else
          \State $\Delta_t \gets C_{\text{ratio}}[t] \cdot \Delta_{t+1}$
          % \State $\Delta_t \gets r \cdot \Delta_{t+1}$ \Comment{second-order delta prediction}
          \State $\hat{\epsilon}_t \gets \hat{\epsilon}_{t+1} + \Delta_t$
        \EndIf
        \State $\mathbf{x}_{t-1} \gets \Call{SchedulerStep}{\mathbf{x}_t, \hat{\epsilon}_t, t}$
      \EndFor
      \State \Return $\mathbf{x}_0$
    \EndFunction
  \end{algorithmic}
\end{algorithm}

\subsection{Orthogonality}

\paragraph{Scope.}
\textsc{TaoCache} operates via output–noise deltas in late denoising stages, which is largely disjoint from spatial–temporal sparsity methods and prior feature-caching that occurs in mid inference steps. It is training-free and applies on top of a given sampler ($G_t$), making it compatible with both cache-based and non-cache accelerations.

As concrete illustrations, we demonstrate one composition along the \emph{timestep} axis (Tea+Tao Cache) and one along the \emph{spatial–temporal} axis (PAB+TaoCache). We did not empirically study other combinations like Delta-DiT or AdaCache; they should be compatible in principle, which we leave for future work.

\subsubsection{Orthogonal: Tea + Tao Cache}
Integrating \textsc{TaoCache} with \textsc{TeaCache} is straightforward: allocate \emph{late} steps to \textsc{TaoCache} and apply \textsc{TeaCache} to the remaining (earlier/mid) steps, with a short \emph{refresh guard band} between the two ranges to prevent error carryover.

\begin{algorithm}[H]
  \caption{Hybrid \textsc{Tea+Tao} Inference}
  \label{alg:tea_tao}
  \begin{algorithmic}[1]
    \Function{HybridForward}{$\mathbf{x}_T, C_{\text{cos}}, N_{\text{TaoSkip}}, \text{RefreshSteps}$}
      \State $\mathcal{T}_{\text{Tao}} \gets \Call{MaxCosSlidingWindow}{C_{\text{cos}}, N_{\text{TaoSkip}}}$ \Comment{contiguous late-stage window}
      \State $t_{\text{brk}} \gets \max(\mathcal{T}_{\text{Tao}}) + \text{RefreshSteps}$ \Comment{2--3 steps recommended}
      \For{$t = T,\,T\!-\!1,\,\dots,\,t_{\text{brk}}$}
        \State \textbf{Apply} \textsc{TeaCache} step
      \EndFor
      \For{$t = t_{\text{brk}},\,t_{\text{brk}}\!-\!1,\,\dots,\,1$}
        \State \textbf{Apply} \textsc{TaoCache} step
      \EndFor
      \State \Return $\mathbf{x}_0$
    \EndFunction
  \end{algorithmic}
\end{algorithm}

% \subsection{Orthorgonal : PAB + Tao Cache}

% The orthogonality of TaoCache towards PAB is straightforward, as these two methods are entirely different in terms of the flops reduction. We simply combine TaoCache and PAB as they are.

\subsection{Orthogonal: PAB + Tao Cache}

\textsc{PAB} reduces FLOPs by reusing multi-scale attention context within each model call, while \textsc{TaoCache} reduces the \emph{number} of model calls via timestep skipping. Since they act on different axes (intra-step vs.\ inter-step), we simply \emph{enable PAB inside each forward} and let \textsc{TaoCache} choose the skip window. No change to either mechanism is required, aside from ensuring that the \textsc{PAB} state (if any) is reinitialized on refresh steps.

% \subsubsection{Orthorgonal : $\Delta$ DiT + Tao Cache}

\section{Experiments \& Results}

\subsection{Experimental Settings}

\paragraph{Base Models and Compared Methods.}  
To validate the generality of TaoCache, we apply it to three representative DiT‐based video generators:
 {Latte-1 2B}~\cite{latte}, 
 {OpenSora-Plan v110}~\cite{opensoraplan}, and {Wan2.1-1.3B}~\cite{wan2.1}.  
We compare against recent training-free caching and acceleration techniques:  
TeaCache~\cite{teacache} and MagCache~\cite{magcache}. 70 prompts that used for evaluation are sampled from CompBench ~\cite{sun2024t2vcompbench}. Unless otherwise stated, we follow each model’s default inference resolution and hold the sampler and guidance settings fixed across methods.

\paragraph{Evaluation Metrics.}  
We measure {inference efficiency} via the number of inference steps skipped (which is also relative FLOPs reduction).  
For {visual quality}, we report:  
LPIPS (Learned Perceptual Image Patch Similarity; lower is better)~\cite{zhang2018unreasonable}, which uses deep network activations to approximate human perceptual judgments;  
SSIM (Structural Similarity Index; higher is better)~\cite{wang2004ssim}, which quantifies image quality based on luminance, contrast, and structural agreement;  
PSNR (Peak Signal-to-Noise Ratio; higher is better)~\cite{hore2010psnr}, which measures pixel-wise fidelity in decibels.

\subsection{Latte-1 2B}

Latte-1 2B is a DiT-based video generator. We compare \textsc{TaoCache} with \textsc{TeaCache} under \emph{matched skip budgets}, keeping the sampler, guidance, and resolution identical to the baseline. The percentage in parentheses denotes the fraction of steps skipped, and speedup can be calculated accordingly. 

\begin{table}[H]
  \centering
  \caption{%
  Video Quality Metrics and End‐to‐End Speedup for Latte-1 2B.}
  \label{tab:latte2b_results}
  \begin{tabular}{lccccc}
    \toprule
    Method   & \% Speedup & LPIPS $\downarrow$ & SSIM $\uparrow$ & PSNR $\uparrow$ \\
             & (with \% inference steps skipped)  &                   &                 & (dB)            \\
    \midrule
    Original & --      & --                & --              & --              \\
    \midrule
    TeaCache & 21.47\% (17.68\%)      & 0.1674      & 0.7507         & 21.57           \\
    TaoCache & {21.95\% (18.00\%)} & \textbf{0.0598} & \textbf{0.8838} & \textbf{29.36}  \\
    \midrule
    TeaCache & 16.27\% (14.00\%)      & 0.1541      & 0.7654         & 22.21           \\
    TaoCache & {16.27\% (14.00\%)} & \textbf{0.0383} & \textbf{0.9194} & \textbf{32.18}  \\
    \midrule
    TeaCache & 11.11\% (10.00\%)      & 0.1338      & 0.7891         & 23.14           \\
    TaoCache & {11.11\% (10.00\%)} & \textbf{0.0184} & \textbf{0.9524} & \textbf{36.07}  \\
    \midrule
    TeaCache &  6.80\% (6.37\%)       & 0.1017      & 0.8334         & 25.46           \\
    TaoCache & {6.38\% (6.00\%)}  & \textbf{0.0127} & \textbf{0.9634} & \textbf{38.03}  \\
    \bottomrule
  \\
  \end{tabular}
  \\
  % \caption{%
  % Video Quality Metrics and End‐to‐End Speedup for Latte-1 2B.\\
  % \textbf{Model:} Latte 2B\\
  \raggedright
  {\textbf{Full Timesteps:} 50\\
  \textbf{Frames:} 16\\
  \textbf{Resolution:} 512\,$\times$\,512\\
  \textbf{Prompts:} 70 in total (7 domains, 10 prompts each)
  }
\end{table}

\begin{figure}[H]
    \centering
    \includegraphics[width=1\linewidth]{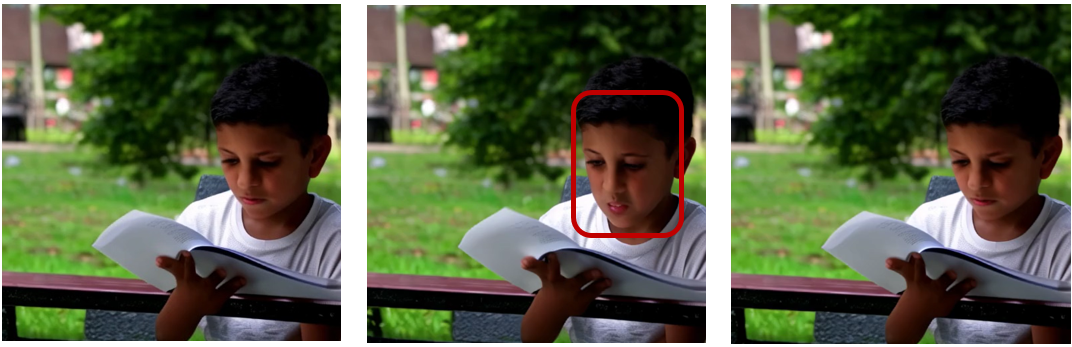}
    \captionsetup{justification=centering}
    \caption{Baseline vs TeaCache vs TaoCache (ours) \\ Latte-1 2B (16 frames), 6\% Inference Steps Skipped}
    \label{fig:enter-label}
\end{figure}

\begin{figure}[H]
    \centering
    \includegraphics[width=1\linewidth]{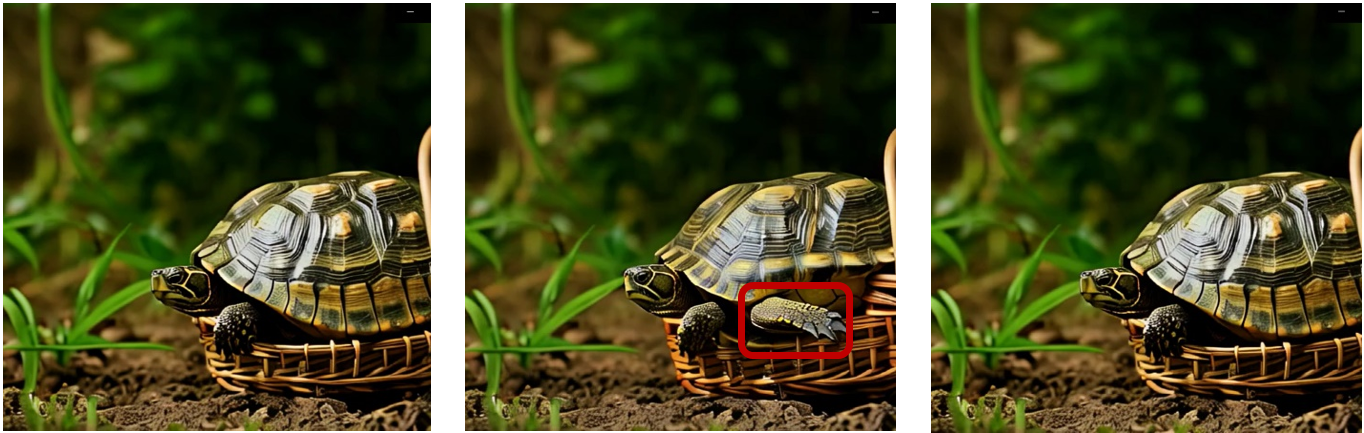}
    \captionsetup{justification=centering}
    \caption{Baseline vs TeaCache vs TaoCache (ours) \\ Latte-1 2B (16 frames), 10\% Inference Steps Skipped}
    \label{fig:enter-label}
\end{figure}

\begin{figure}[H]
    \centering
    \includegraphics[width=1\linewidth]{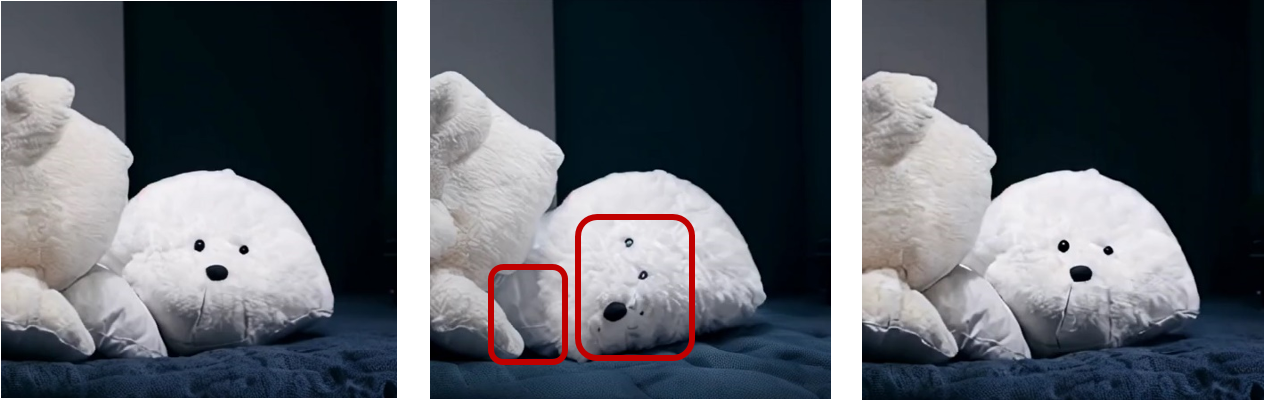}
    \captionsetup{justification=centering}
    \caption{Baseline vs TeaCache vs TaoCache (ours) \\ Latte-1 2B (16 frames), 18\% Inference Steps Skipped}
    \label{fig:enter-label}
\end{figure}

These are three sample comparisons across different scales (6\%, 10\%, 18\%) of inference step skips on Latte-1 2B. From the results, those undesired caching behaviours like disfigured faces and unexpected limbs are avoided by TaoCache. This is significant for high-standard video generations of instruction following for controller and facial consistency for story alignment. 

We should emphasize that not all TeaCache's generations would have these problems, but they are the issues that TeaCache cannot inherently avoid.

\subsection{OpenSora-Plan v110}

OpenSora-Plan is an open-source DiT-based video generation framework that targets high-fidelity and efficient synthesis at moderate resolutions. Similar to the Latte-1 case, we evaluate \textsc{TaoCache} against \textsc{TeaCache} under matched skip budgets, keeping the sampling algorithm, guidance scale, and resolution fixed. \textsc{TaoCache} uses its late-stage, deviation-aware skip placement to improve stability and quality, particularly for temporally consistent content and fine details.

\begin{table}[H]
  \centering
    \caption{%
    Video Quality Metrics and End‐to‐End Speedup for OpenSora-Plan v110.\\}
  \label{tab:opensoraplan_v110}
  \begin{tabular}{lccccc}
    \toprule
    Method   & \% Speedup      & LPIPS $\downarrow$ & SSIM $\uparrow$ & PSNR $\uparrow$ \\
             &  (with \% inference steps skipped)    &                    &                 & (dB)            \\
    \midrule
    Original & --              & --                 & --              & --              \\
    \midrule
    TeaCache & 33.68\% (25.2\%) & 0.0780            & 0.8531          & 26.27           \\
    TaoCache & {33.33\% (25.0\%)} & \textbf{0.0472} & \textbf{0.8873} & \textbf{29.71}  \\
    \midrule
    TeaCache & 26.93\% (21.2\%) & 0.0774            & 0.8538          & 26.32           \\
    TaoCache & {26.58\% (21.0\%)} & \textbf{0.0330} & \textbf{0.9093} & \textbf{31.58}  \\
    \midrule
    TeaCache & 19.47\% (16.3\%) & 0.0586            & 0.8842          & 28.40           \\
    TaoCache & {19.04\% (16.0\%)} & \textbf{0.0294} & \textbf{0.9151} & \textbf{32.24}  \\
    \midrule
    TeaCache &  8.69\% (8.0\%)  & 0.0512            & 0.8968          & 29.13           \\
    TaoCache & {9.89\% (9.0\%)}  & \textbf{0.0164} & \textbf{0.9485} & \textbf{35.40}  \\
    \bottomrule
    \\
  \end{tabular}
  \\ 
  \raggedright
    {% \textbf{Model:} OpenSora-Plan v110\\
    \textbf{Full Timesteps:} 100 (9 scheduler order + 91)\\
    \textbf{Frames:} 65\\
    \textbf{Resolution:} 512\,$\times$\,512\\
    \textbf{Prompts:} 70 in total (7 domains, 10 prompts each)
  }
\end{table}

\begin{figure}[H]
    \centering
    \includegraphics[width=1.01\linewidth]{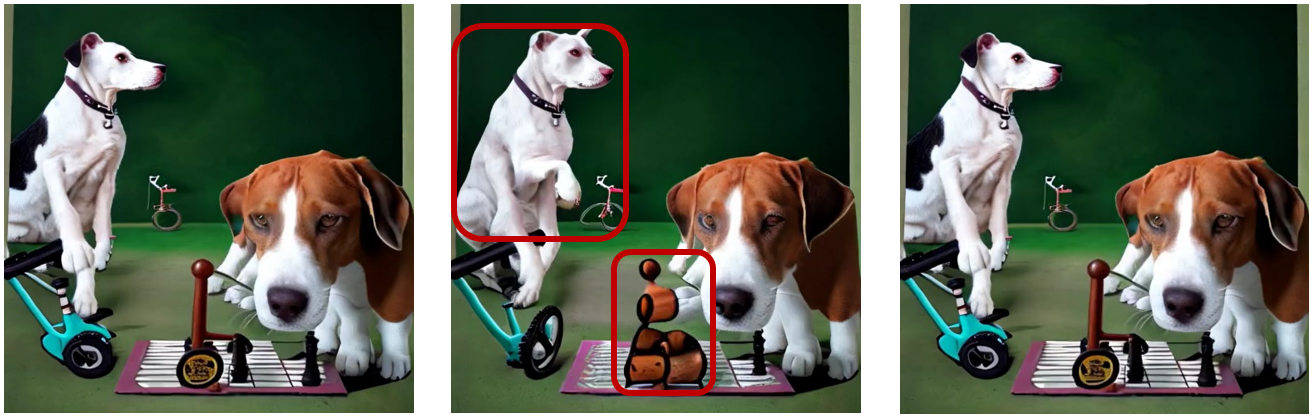}
    \captionsetup{justification=centering}
    \caption{Baseline vs TeaCache (8\%) vs TaoCache (ours) \\ OpenSora-Plan v110 (65 frames), 9\% Inference Steps Skipped}
    \label{fig:enter-label}
\end{figure}

\begin{figure}[H]
    \centering
    \includegraphics[width=1.01\linewidth]{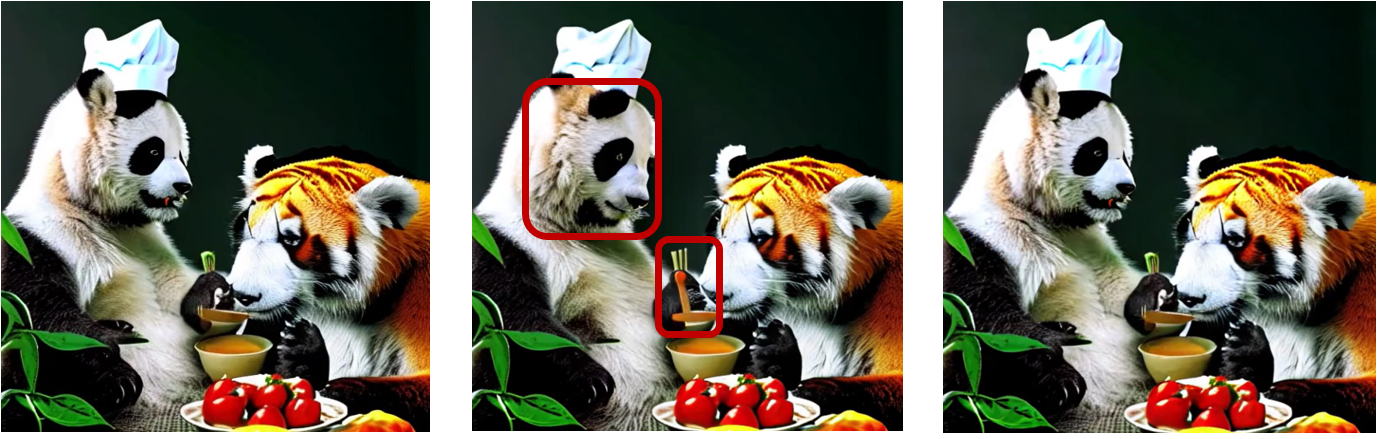}
    \captionsetup{justification=centering}
    \caption{Baseline vs TeaCache vs TaoCache (ours) \\  OpenSora-Plan v110 (65 frames), 16\% Inference Steps Skipped}
    \label{fig:enter-label}
\end{figure}

\begin{figure}[H]
    \centering
    \includegraphics[width=1.01\linewidth]{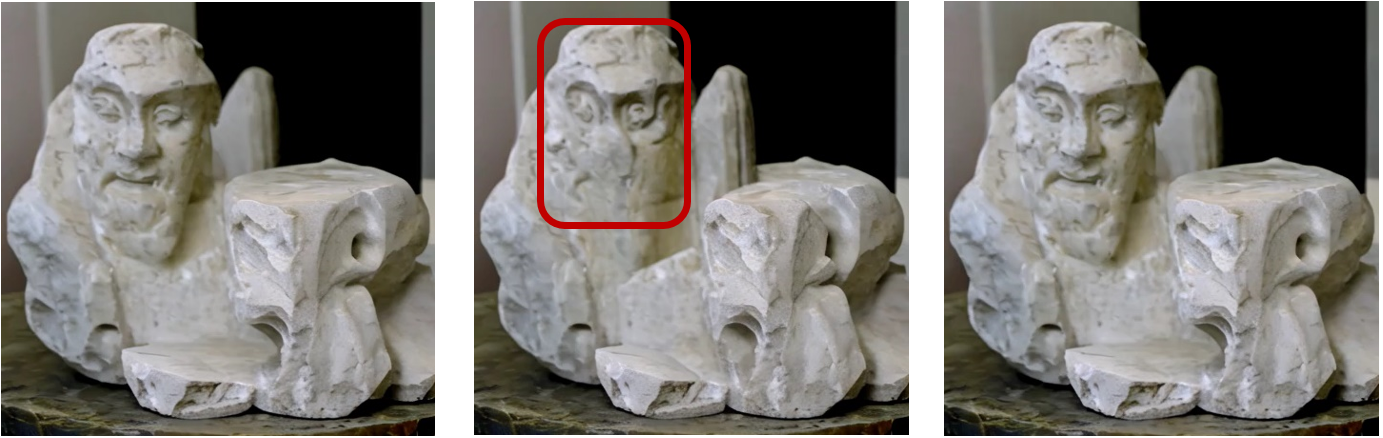}
    \captionsetup{justification=centering}
    \caption{Baseline vs TeaCache vs TaoCache (ours) \\  OpenSora-Plan v110 (65 frames), 21\% Inference Steps Skipped}
    \label{fig:enter-label}
\end{figure}

For OpenSora-Plan, TaoCache has the same expected behaviour as we discussed in Latte-1 2B's experiments. For TeaCache of 8.0\% step skips, these 8 skips (since 100 timesteps in total) come from the very early stages of warmup timesteps and cannot easily be further reduced to 4 skips by tuning the {\it $rel\_l1\_thresh$} parameter.

\subsection{Wan2.1 1.3B}

Wan2.1-1.3B is a state-of-the-art DiT-based video generation model designed for high-fidelity, controllable synthesis. 
In addition to \textsc{TeaCache}, we also compare against \textsc{MagCache}~\cite{magcache} on this model.
All methods use the model’s default inference settings (sampler, guidance, resolution) to ensure comparability. 
Skip budgets are matched across methods; speedup is reported with parentheses indicating the proportion of denoising steps skipped. 
\textsc{TaoCache} applies its calibrated, deviation-aware skip placement in late-stage timesteps to maximize stability and visual quality.

\begin{figure}[H]
  \centering
  \includegraphics[width=1.02\linewidth]{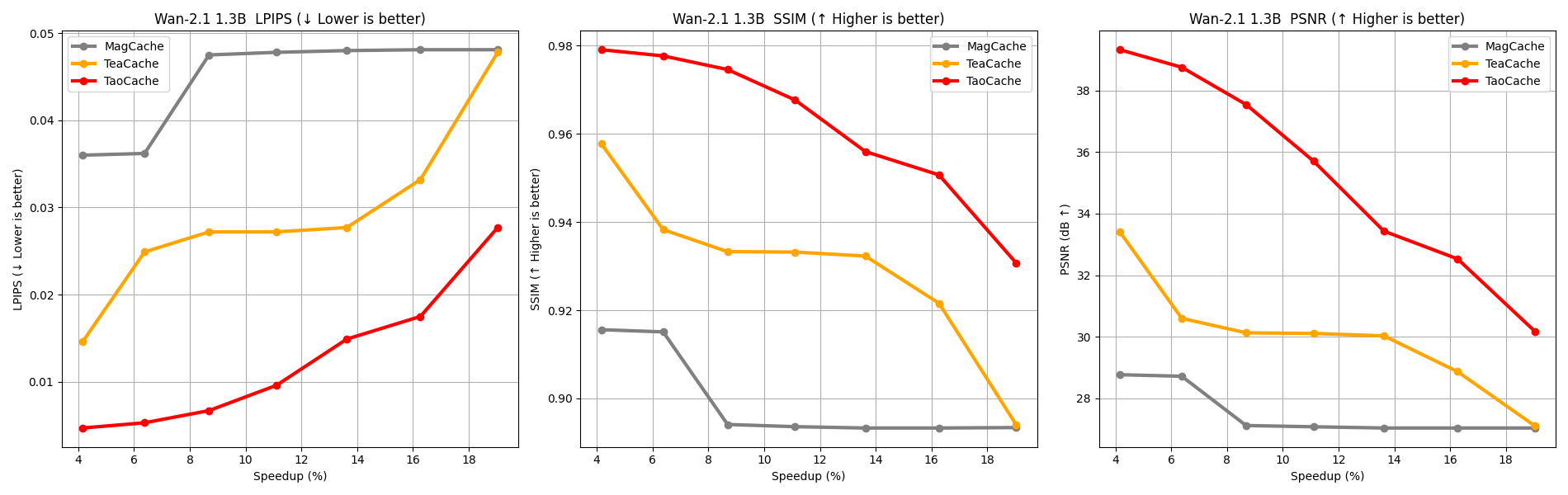}
  \caption{%
    Comparisons between TaoCache, TeaCache, and MagCache on Wan2.1-1.3B.\\
    % \textbf{Model:} Wan2.1-1.3B\\
    \textbf{Full Timesteps:} 50\\
    \textbf{Frames:} 33 (2\,s)\\
    \textbf{Resolution:} 832\,$\times$\,480 (480\,p)\\
    \textbf{Prompts:} 70 in total (7 domains, 10 prompts each)
  }
  \label{fig:video-comparison}
\end{figure}

% First figure, one per line
\begin{figure}[H]
  \centering
  \includegraphics[width=1.02\linewidth]{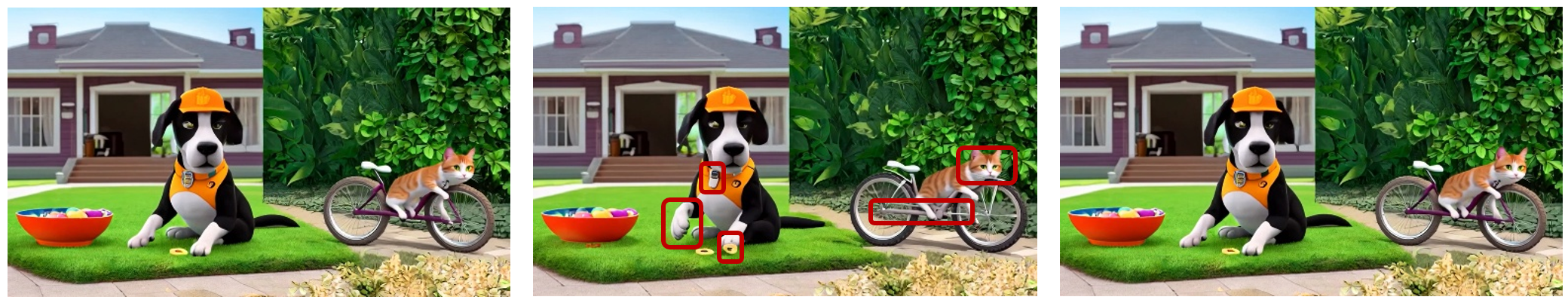}
  \captionsetup{justification=centering}
  \caption{Baseline vs TeaCache vs TaoCache (ours) \\  Wan2.1-1.3B (33 frames), 8\% Inference Steps Skipped}
  \label{fig:8pct_5_0}
\end{figure}

% Second figure
\begin{figure}[H]
  \centering
  \includegraphics[width=1.02\linewidth]{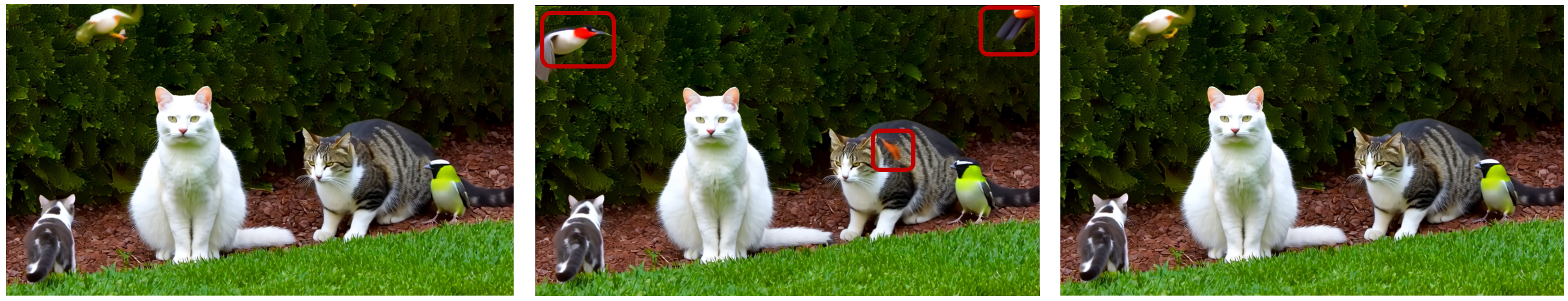}
  \captionsetup{justification=centering}
  \caption{Baseline vs TeaCache vs TaoCache (ours) \\  Wan2.1-1.3B (33 frames), 8\% Inference Steps Skipped}
  \label{fig:8pct_7_1}
\end{figure}

% Third figure
\begin{figure}[H]
  \centering
  \includegraphics[width=1.02\linewidth]{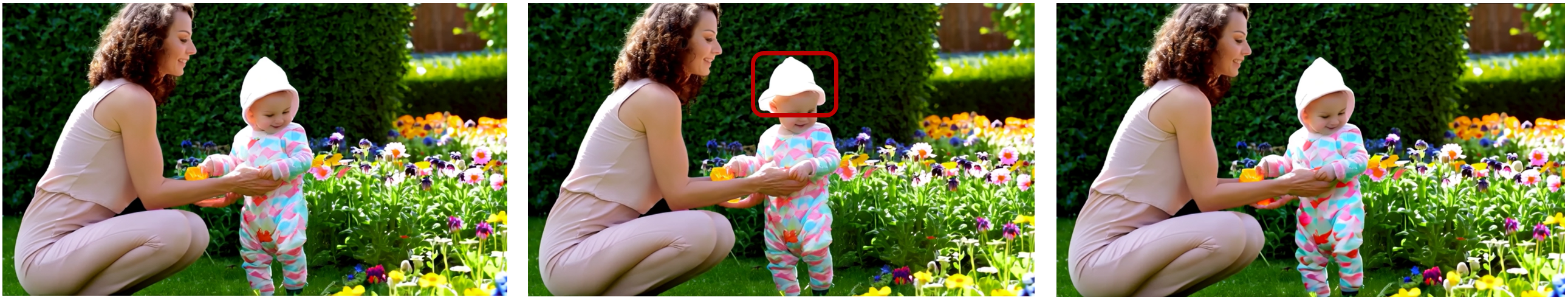}
  \captionsetup{justification=centering}
  \caption{Baseline vs TeaCache vs TaoCache (ours) \\  Wan2.1-1.3B (33 frames), 16\% Inference Steps Skipped}
  \label{fig:16pct_1_6}
\end{figure}

% Fourth figure
\begin{figure}[H]
  \centering
  \includegraphics[width=1.02\linewidth]{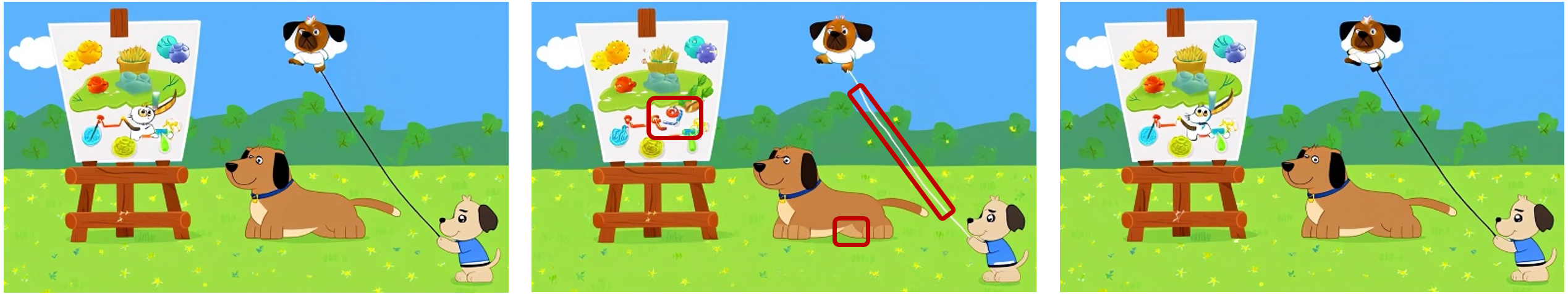}
  \captionsetup{justification=centering}
  \caption{Baseline vs TeaCache vs TaoCache (ours) \\ Wan2.1-1.3B (33 frames), 16\% Inference Steps Skipped}
  \label{fig:16pct_5_2}
\end{figure}

These are the sample result from TaoCache and TeaCache on Wan2.1. Figure~\ref{fig:video-comparison} shows that TaoCache surpasses the other two methods under 20\% speedup. However, from Figure~\ref{fig:16pct_1_6} we can inspect that the TaoCache strategy also have certain flaws. The color of the baby suit is denser in TaoCache. This can be explained as TaoCache may overestimate the denoising momentum for a certain prompt in a long skip of late denoising stages.

\subsection{Ablation for Caching Feature}

To compare the residual-based caching and output-delta caching in the later denoising process, we test TeaCache's residual mechanism on the same timesteps being skipped by TaoCache. This is indicated as TaoSkip + TeaResidual rows in the following table.  

From the following table, we see that, on the same timesteps skipping in late denoising procedures, the residual skipping strategy is not as good as the output delta skipping.

\begin{table}[H]
  \centering

  \label{tab:wan21_3b_results}
  \begin{tabular}{lccccc}
    \toprule
    Method            & \% Speedup                      & LPIPS $\downarrow$ & SSIM $\uparrow$ & PSNR $\uparrow$ \\
                      & (with \% inference steps skipped) &                   &                 & (dB)            \\
    \midrule
    Original          & --                              & --                & --              & --              \\
    \midrule
    TeaCache          & 4.16\% (4.00\%)                 & 0.0146            & 0.9578          & 33.41           \\
    TaoSkip+TeaResidual    & 4.16\% (4.00\%)                 & 0.0088            & 0.9704          & 36.63           \\
    TaoCache          & 4.16\% (4.00\%)                 &  \textbf{0.0047}            &  \textbf{0.9791}          &  \textbf{39.32}           \\
    \midrule
    TeaCache          & 8.69\% (8.00\%)                 & 0.0272            & 0.9333          & 30.13           \\
    TaoSkip+TeaResidual    & 8.69\% (8.00\%)                 & 0.0126            & 0.9633          & 35.19           \\
    TaoCache          & 8.69\% (8.00\%)                 &  \textbf{0.0067}            &  \textbf{0.9746}          &  \textbf{37.54}           \\
    \midrule
    TeaCache          & 13.63\% (12.00\%)               & 0.0277            & 0.9323          & 30.03           \\
    TaoSkip+TeaResidual    & 13.63\% (12.00\%)               & 0.0215            & 0.9465          & 33.02           \\
    TaoCache          & 13.63\% (12.00\%)               &  \textbf{0.0149}            &  \textbf{0.9560}          &  \textbf{33.43}           \\
    \midrule
    TeaCache          & 19.04\% (16.00\%)               & 0.0478            & 0.8940          & 27.11           \\
    TaoSkip+TeaResidual    & 19.04\% (16.00\%)               & 0.0333            & 0.9200          &  \textbf{30.87}           \\
    TaoCache          & 19.04\% (16.00\%)               &  \textbf{0.0277}            &  \textbf{0.9307}          & 30.18           \\
    \bottomrule
  \\
  \end{tabular}
  \\
  \caption{%
  Video Quality Metrics and End‐to‐End Speedup for Wan2.1-1.3B.\\
  % \textbf{Model:} Wan2.1-1.3B\\
  \textbf{Full Timesteps:} 50\\
  \textbf{Frames:} 33 (2s)\\
  \textbf{Resolution:} 832\,$\times$\,480 (480p)\\
  \textbf{Prompts:} 70 in total (7 domains, 10 prompts each) }
\end{table}

\subsection{Orthorgonality : Tea + Tao Cache}

\begin{figure}[H]
    \centering
    \includegraphics[width=1\linewidth]{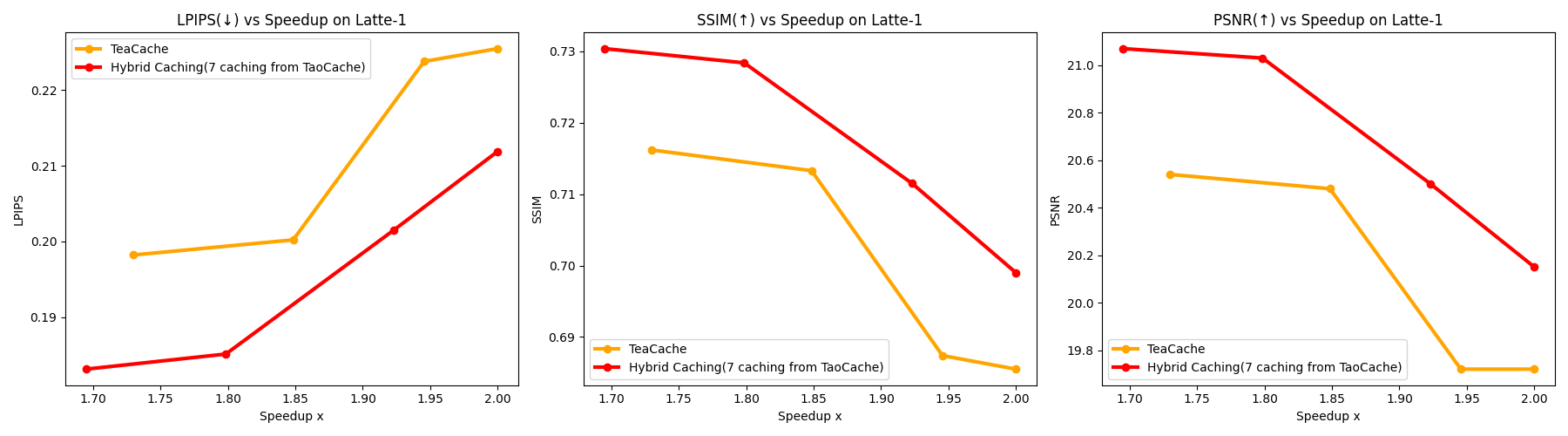}
    \caption{Hybrid Caching(TeaCache + TaoCache) on Latte-1 2B \\
    % \textbf{Model:} Latte-1 2B\\
    \textbf{Full Timesteps:} 50\\
    \textbf{Frames:} 16 \\
    \textbf{Resolution:} 512\,$\times$\,512 \\
    \textbf{Prompts:} 70 in total (7 domains, 10 prompts each)}
    \label{fig:orth_teatao}
\end{figure}

Mentioned in Algorithm ~\ref{alg:tea_tao}, the Hybrid Cache firstly generates under TeaCache's range and then ends with TaoCache's inference caching. The experiments in Fig~\ref{fig:orth_teatao} are conducted between pure TeaCache and Hybrid Caching with the TeaCache followed by 7 steps of timestep skips from TaoCache. 

The results show that with the same percentages of acceleration, the Hybrid Cache can surpass the pure TeaCache method in terms of video quality.

\subsection{Orthorgonal : PAB + TaoCache}

The experiment for PAB with TaoCache is conducted directly based on PAB$_{224}$, PAB$_{236}$, PAB$_{347}$, PAB$_{469}$ settings, where PAB$_{\alpha \beta \gamma}$ represents the broadcast ranges of spatial ($\alpha$), temporal ($\beta$), and cross ($\gamma$) attentions~\cite{latte}.

The following graphs show that with comparable video qualities, PAB + TaoCache can significantly speed up the inference.
\begin{figure}[H]
    \centering
    \includegraphics[width=1\linewidth]{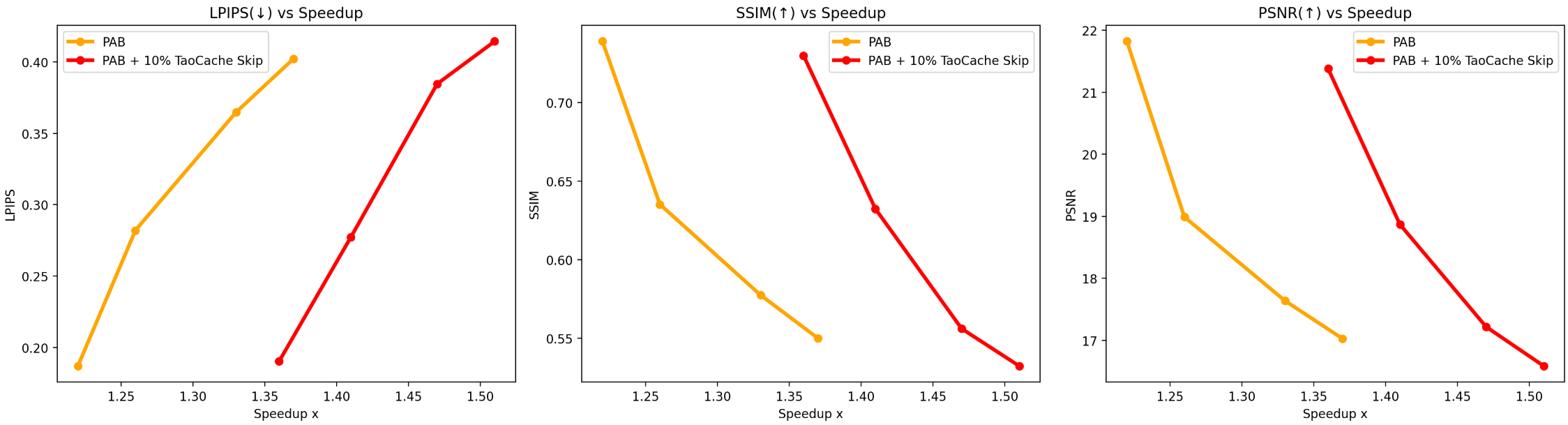}
    \caption{
      PAB + TaoCache on Latte-1 2B.\\
      % \textbf{Model:} Latte 2B\\
      \textbf{Full Timesteps:} 50\\
      \textbf{Frames:} 16\\
      \textbf{Resolution:} 512\,$\times$\,512\\
      \textbf{Prompts:} 70 in total (7 domains, 10 prompts each) }
    \label{fig:placeholder}
\end{figure}

% \subsection{Orthorgonal : $\Delta$ DiT + Tao Cache}

\section{Limitations}

There are three main constraints for TaoCache. Firstly, the inference steps range that TaoCache can be applied is narrower than TeaCache and MagCache, as it only applies in late stages, but its orthogonality compensates for this. Secondly, the calibration is as heavy as TeaCache. To inspect the deviation of output delta's cosine similarities and norm ratio, for a new model, 20 prompts from various domains are recommended. Lastly, for a model trained under uniformly distributed timesteps~\cite{ho2020ddpm,peebles2023dit}, late-stage behavior is relatively predictable, making second-order delta prediction effective. In contrast, for models trained with \emph{log-normal} timestep sampling~\cite{opensora2024,flux2025}, more structural updates concentrate at late steps; the deltas become less stationary, and \textsc{TaoCache} may be less effective.

\section{Conclusion}

In this work, we introduced TaoCache, a novel training-free acceleration method for diffusion-based video generation that prioritizes the preservation of structural integrity. We identified that existing caching methods, which primarily skip early or middle denoising steps, can lead to discrepancies in the final video's composition and character consistency. To address this, TaoCache introduces a caching strategy based on a fixed-point approximation of the second-order noise delta. By leveraging the observation that this delta becomes highly stable and predictable during the late stages of denoising, our method effectively skips computations where it matters least for structure and most for fine-tuning details.

Our extensive experiments on state-of-the-art models like Latte, OpenSora-Plan, and Wan2.1 demonstrate that TaoCache significantly outperforms prior caching techniques such as TeaCache and MagCache, achieving superior visual quality across LPIPS, SSIM, and PSNR metrics for the same level of speedup. We also showed that TaoCache is orthogonal to other acceleration methods and can be successfully hybridized to yield even greater efficiency. While its effectiveness is currently focused on models trained with uniform timestep distributions, TaoCache presents a robust and principled approach to accelerating video generation while maintaining high fidelity and structural coherence, paving the way for more practical applications of large-scale video diffusion models.

\newpage

\appendix

\section{Appendix}

\subsection{Model Calibrations}

\begin{figure}[H]
    \centering
    \includegraphics[width=1\linewidth]{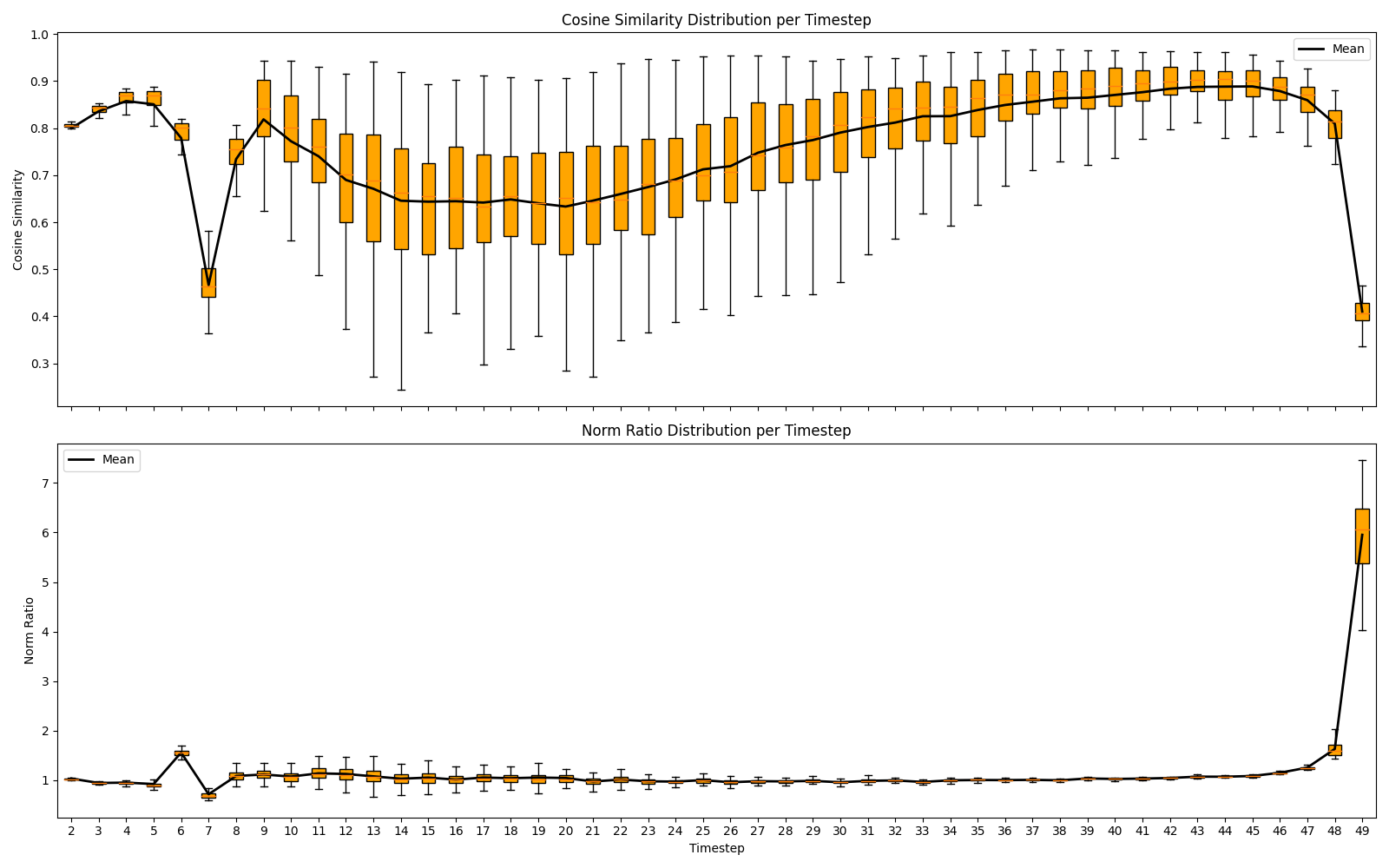}
    \caption{Latte-1 2B (16 frames)}
    \label{fig:enter-label}
\end{figure}
\begin{figure}[H]
    \centering
    \includegraphics[width=1\linewidth]{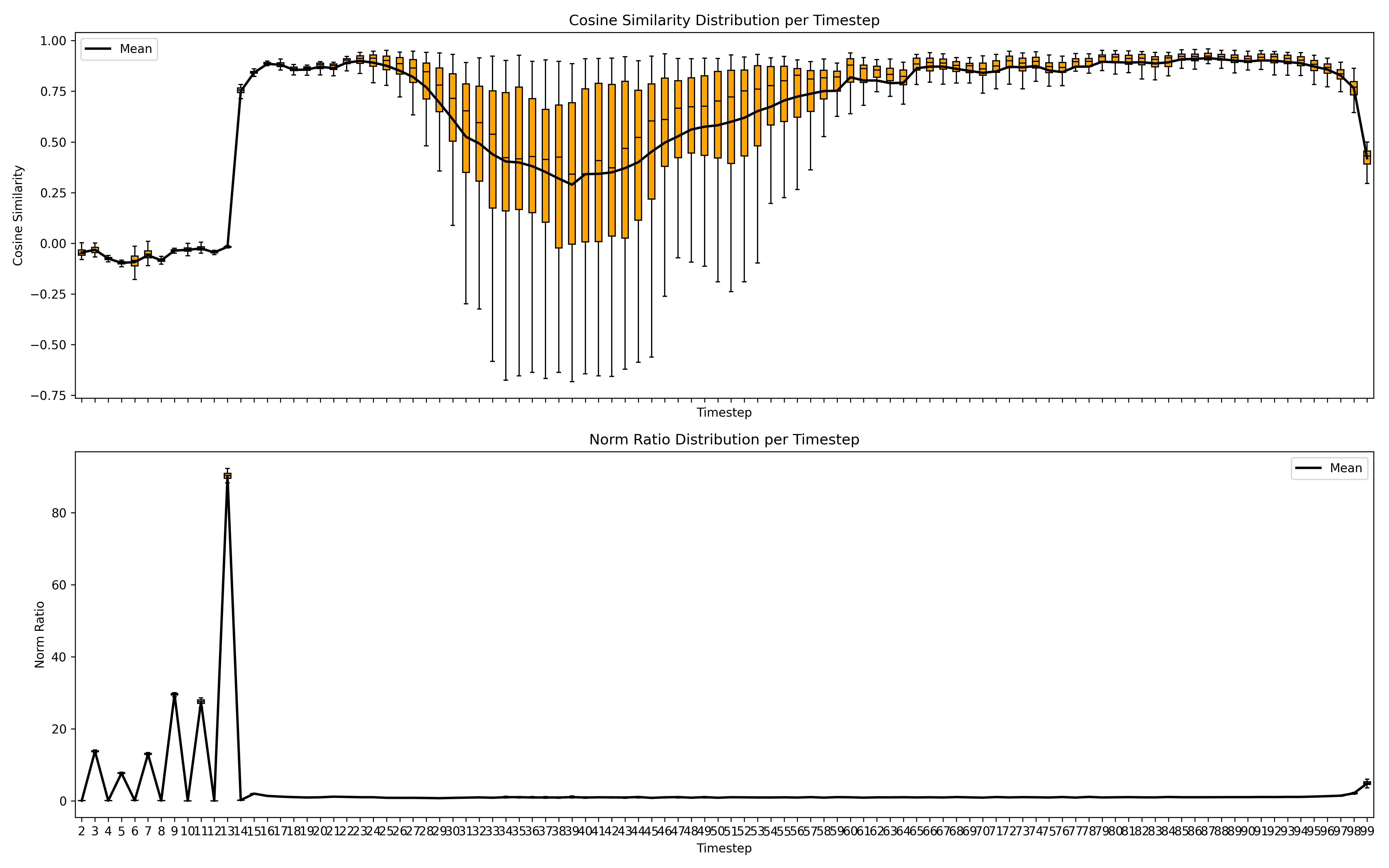}
    \caption{Opensora-Plan V110 (65 frames)}
    \label{fig:enter-label}
\end{figure}

% \begin{ack}

% \end{ack}

%%%%%%%%%%%%%%%%%%%%%%%%%%%%%%%%%%%%%%%%%%%%%%%%%%%%%%%%%%%%

% \appendix

% \section{Appendix / supplemental material}

% Optionally include supplemental material (complete proofs, additional experiments and plots) in appendix.
% All such materials \textbf{SHOULD be included in the main submission.}

%%%%%%%%%%%%%%%%%%%%%%%%%%%%%%%%%%%%%%%%%%%%%%%%%%%%%%%%%%%%

% \newpage
% \section*{NeurIPS Paper Checklist}

\end{document}